\documentclass[runningheads]{llncs}
\usepackage[T1]{fontenc}
\usepackage{graphicx}
\usepackage{hyperref}
\usepackage{float}
\begin{document}
\title{Vibrotactile information coding strategies for a body-worn vest to aid robot-human collaboration\thanks{This Paper is a Preprint version made available on Arxiv 28/02/2025}}
\titlerunning{Comparison of vibrotactile information coding strategies}

\author{Adrian Vecina Tercero\inst{1} \and
Praminda Caleb-Solly\inst{1}}
\authorrunning{A. Vecina Tercero, P. Caleb-Solly}
%
\institute{School of Computer Science, University of Nottingham, Nottingham, UK \\
%
\email{\{psyav2,pszpc1\}@nottingham.ac.uk}}
\maketitle      
\begin{abstract}
This paper explores the use of a body-worn vibrotactile vest to convey real-time information from robot to operator.   
Vibrotactile communication could be useful in providing information without compromising or loading a person's visual or auditory perception. This paper considers applications in Urban Search and Rescue (USAR) scenarios where a human working alongside a robot is likely to be operating in high cognitive load conditions.
The focus is on understanding how best to convey information considering different vibrotactile information coding strategies to enhance scene understanding in scenarios where a robot might be operating remotely as a scout. In exploring information representation, this paper introduces the idea of Semantic Haptics, using shapes and patterns to represent certain events as if the skin was a screen, and shows how these lead to better learnability and interpretation accuracy.

\end{abstract}
\section{Introduction}
As robotic technologies progress, the potential applications also expand. One service application area garnering attention is Urban Search and Rescue (USAR). USAR Robots are designed to navigate hazardous environments and augment human capabilities in rescue operations. However, effective collaboration between human rescue operators and these robots requires overcoming challenges in human-robot interaction related to the complexity of dynamic and unpredictable disaster scenarios. Urban disaster environments can present various risks to human rescue operators, hazards such as extreme temperatures, asbestos, dust, toxic gases, structural collapse and more ~\cite{casper_2003_humanrobot}. While employing robots can mitigate the risks that human rescuers have to take, controlling them in such unpredictable environments is likely to be cognitively demanding. 

This paper presents a study on assessing vibrotactile feedback strategies, with the aim of  evaluating the interpretability of information received from the robot with minimal training. The intended future impact of this approach is to enhance human-robot collaboration (HRC) scenarios requiring real-time feedback by providing a reliable communication channel that does not compromise the main visual and auditory senses of the human operator.

\section{Related Work}

\subsection{Vibrotactile Haptics}

The human skin is highly sensitive, able to discern small changes in touch, pressure, texture, and temperature \cite{johansson_1978_tactile}. This sensitivity could be advantageous for conveying information when visual and auditory channels are occupied or restricted, such as by personal protective equipment or environmental conditions. Existing haptic technologies include vibrotactile, force-feedback, surface display, and tactile displays \cite{hayward_2007_do}. Vibrotactile feedback (also known as vibrohaptics) is the most studied \cite{hayward_2007_do} and is used in products like phones, gaming controllers, and surgical instruments \cite{jones_2008_tactile}. It utilises vibrations of varying duration, intensity, or frequency through eccentric rotating mass (ERM) motors \cite{precisionmicrodrives_2021_eccentric}, which are typically in indirect contact with the skin \cite{hayward_2007_do}. 

\subsection{Vibrotactile Feedback in Robotics}

Vibrotactile feedback has been used for communication in various contexts. In \cite{prasad_2014_haptimoto}, vibrohaptic pads worn on the back indicated distance and direction of approaching turns for a motorcycle rider, using pulses at varying rates and intensities. The study found that users could understand these messages even while riding. In other studies haptics have been used as a form of force-feedback to assist in teleoperation of robots \cite{tri_2020_mmi,wang_2016_designing,rognon_2019_haptic,elrassi_2020_a}, or as a way to provide information from an autonomous robot \cite{scheggi_2014_humanrobot,scheggi_2012_vibrotactile,bolarinwa_2019_the}. Haptics are also beginning to be used in surgical robotics \cite{elrassi_2020_a,selim_2023_a}. These studies provide important findings in how haptics can be used effectively, and how users work with, and react to, haptics whilst working with robots \cite{haruna_2021_comparison}. Haptics can also enhance human-robot collaboration. In \cite{casalino_2018_operator}, a vibrotactile ring was use to convey that a robot understood the intent of the user in an assembly task. The research found that this enabled faster assembly through better human-robot understanding.

\subsection{Vibrotactile Message Modalities}
This section describes different vibrotactile coding strategies identified.

\noindent \textbf{Simple Activation:} This basic modality consists of applying a vibration for a certain duration and intensity. This is common in game controllers and phones. This strategy allows simple time-based patterns to be built, however these are often used just to distinguish different alerts, not to encode information.

\noindent \textbf{Position Based:} This modality uses multiple motors, and is common in related work \cite{scheggi_2012_vibrotactile,scheggi_2014_humanrobot,casalino_2018_operator,prasad_2014_haptimoto}. Vibrating motors are placed such that they are individually discernible, and the user memorises the alert that each one corresponds to. Motor intensity can be used to represent analogue values. For example, a motor representing robot battery can vibrate more intensely the lower the battery gets. However, past research has found that human skin struggles to detect small changes in tactile intensity \cite{brewster_tactons,jones_2008_tactile}. To overcome this, pulses are used instead \cite{prasad_2014_haptimoto}, as it is easier to discern between vibrating and not vibrating, than between different intensities \cite{brewster_tactons}. This idea is employed later in the directional vibrations.

\noindent \textbf{Pattern Based Messages:} Enabled by a large matrix of vibration motors, this modality uses learnt patterns (rather than locations) representing certain events in the environment (e.g. spirals, zig-zags). These patterns can vary in size, intensity and speed to modify the meaning of the message, such as increased urgency or proximity. 

\noindent \textbf{Semantic Messages:} We extend the above approach by using haptic messages that directly represent what they refer to. For example, a pulsing heart pattern can inform that a person has been detected. Little research exists on this strategy. We hypothesise that associating the shape of the pattern directly to the event could improve learnability and recognition of the information.

\subsection{In the Context of this study}

Most vibrotactile studies use small devices, with simple binary vibrations \cite{scheggi_2012_vibrotactile,scheggi_2014_humanrobot,casalino_2018_operator}, usually in the form of wristbands or rings. These studies report that haptic feedback is generally understood by users, and can be interpreted without extensive training using few motors, and simple messages. In this research, we explore a larger matrix of motors, allowing for more types of information to be conveyed at the same time, using different coding strategies. While this enables more complex messages and information to be transmitted, cognitive load and retention are critical considerations.

A key paradigm explored in this paper is the \emph{semantic coding of vibrations}. We create cutaneous shapes by alternating the patterns of the motors being actuated, so that messages transmitted can take the form of a variety of things being represented. It is hypothesised that these semantic vibrations could allow a user to better associate the message to the concept, enabling better learnability.

\section{Methods}

The primary aim of this study is to investigate the use of a wearable vibrohaptic vest with a large matrix of motors to transmit real-time information from an autonomous robot in USAR-like missions. Whilst USAR is used as an exemplar scenario, the findings could apply to other situations where auditory and visual channels are occupied, and the user lacks line of sight to the robot.

\begin{figure}
    \centering
    \begin{minipage}[b]{0.48\textwidth}
        \centering
        \includegraphics[width=0.9\textwidth]{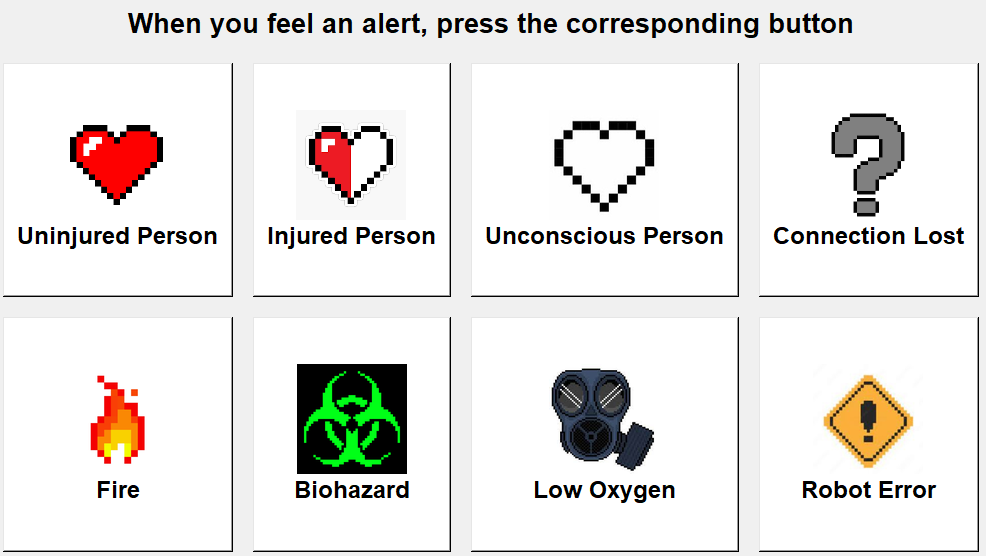}
        \caption{UI used for training and testing}
        \label{fig:ui}
    \end{minipage}
    \hspace{0.02\textwidth}
    \begin{minipage}[b]{0.48\textwidth}
        \centering
        \includegraphics[width=0.8\textwidth]{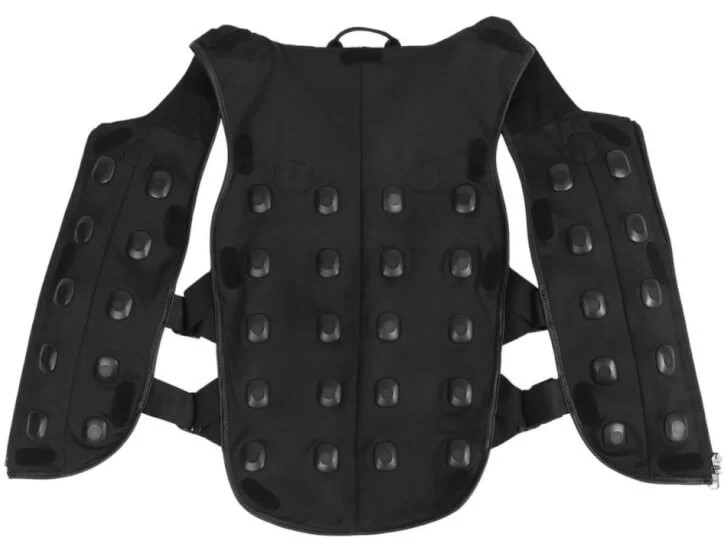}
        \caption{TactVest X40, motors exposed}
        \label{fig:vest}
    \end{minipage}
\end{figure}

In this study, data from camera sensors on a Boston Dynamics Spot robot \cite{bostondynamics_2023_spot}, were processed and presented as vibrotactile "messages" on a bHaptics Tact x40 Haptic Vest \cite{bhaptics_next} (seen in figure \ref{fig:vest}). This wearable vest, originally designed for virtual reality gaming, consists of two 4 by 5 motor matrices, front and back. Each motor can be individually actuated to vibrate at a given intensity.

Another aspect of this study is the design of \emph{vibrotactile patterns}, which should allow users to be immediately aware of the conveyed environmental events. Experiments were conducted with participants wearing the vest to assess the viability of the system, compare different vibrotactile "messages", and evaluate the effects of adding additional mental workload by getting them to perform other cognitive tasks. \cite{Orru}

\noindent \textbf{Event Patterns:} These refer to one-time vibrotactile patterns when the robot detects something in the environment. For this study, 8 example USAR events were chosen. Each had a semantic pattern designed found in table \ref{tab:data3}. These were compared against a simple positional pattern consisting of a 2x2 square vibration in different locations of the vest.

\vspace{-12pt}
\begin{table*}
    \caption{Description of Semantic patterns used in trials.}
    \begin{tabular}{c p{0.75 \textwidth}}
        \hline
        \textbf{Event/Detection} & \textbf{Pattern Description}\\
        \hline
        Uninjured Person & Slow, consistent heartbeat-like pattern with 4 chest area motors \\
        Injured Person & Faster heartbeat followed by a zig-zag travelling across the chest \\
        Unconscious Person & Faster heartbeat followed by an explosion-like on the front\\
        Fire &  Slow wave that travels upwards from the stomach to the neck \\
        Low Oxygen & Fast-travelling spiral starting at the lung area. Repeats twice \\
        Bio-hazard &  Two static chevron shapes which vibrate intensely \\
        Robot Error &  A wave that expands from the centre of the body, then contracts \\
        Connection Lost & A slow vibration that wraps around the lower body twice \\
        \hline
    \end{tabular}
    \label{tab:data3}
\end{table*}

\vspace{-12pt}

\begin{figure}
    \centering
    \begin{minipage}{0.4\textwidth}
        \centering
        \includegraphics[width=0.9 \textwidth]{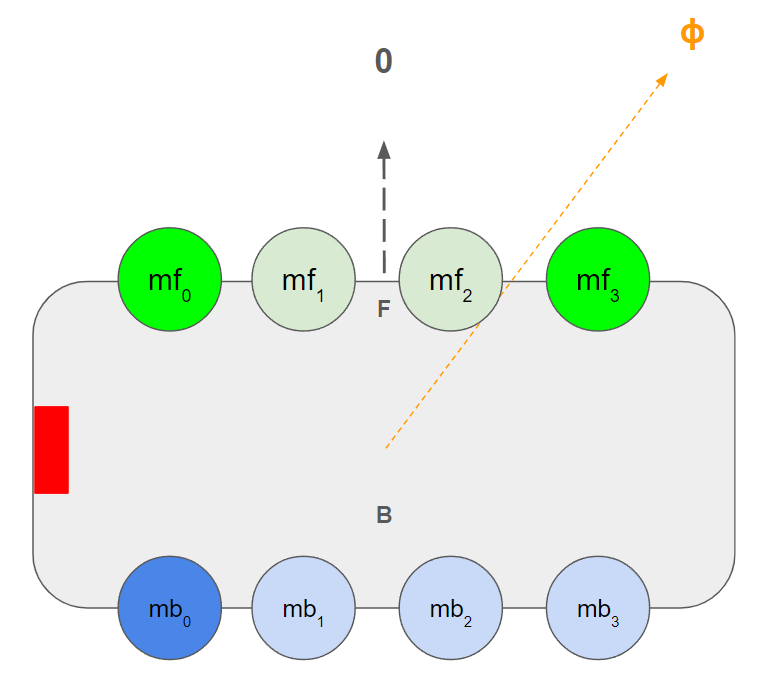}
        \caption{Top-Down representation of the band of motors used for directions. Motors $mf_0, mf_3, mb_0$ are active}. 
        \label{fig:label}
    \end{minipage}
    \hspace{0.02\textwidth}
    \begin{minipage}{0.48\textwidth}
        \centering
    
        \includegraphics[width=1.2 \textwidth]{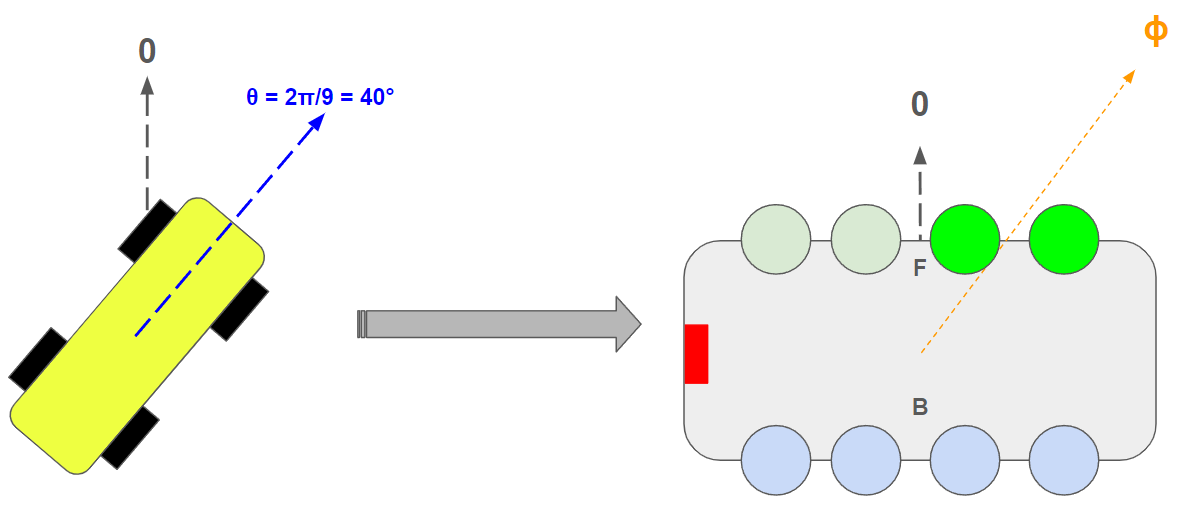}
        \caption{Top-Down representation of Spot Robot, along with displayed haptic direction}
        \label{fig:dir1}
    \end{minipage}
\end{figure}

\noindent \textbf{Direction and Movement:} Within our USAR scenario, effective human-robot collaboration requires that the rescuers are aware of the robot's position. However, it is likely that the robot will not be in view of the rescuer. To overcome this, we employ the vest to provide rudimentary positional data. This is achieved by using a band of 8 motors around the stomach. These motors were chosen after initial user testing found these to be comfortable, and easy to interpret. This band consists of 4 motors in the front and 4 in the back. Two adjacent motors are used to indicate direction, pulsing slowly when the robot was moving, and rapidly when static. The motors used depended on the direction the robot was facing, which was calculated using the robot's odometry data. In this implementation, it is assumed that the direction the user is facing is not known, so the direction of the robot is displayed relative to a calibrated north, shared by the robot and the user wearing the vest. Figure \ref{fig:label} and \ref{fig:dir1} show the band of motors as coloured circles. Also shown is a grey line representing the fixed north $(0)$, and an orange line representing the 'perceived' direction $\phi$ that the user 'feels'. Motors that are in active use are shown with darker colours. This method gives 8 displayable directions. This approach was found to be more effective than using a single motor in early testing. 

\vspace{-15pt}

\section{Evaluation Methods}

Initial testing determined the semantic patterns. We found that if the user is distracted, the information was often missed. To remedy this, we introduced a short alert of three pulses to the region at the base of the neck, which was delivered before any event pattern. The main experiment consisted of two main trials. After these, each participant completed a short questionnaire and gave general feedback. The questions are presented in table \ref{tab:data4} alongside results.

\noindent \textbf{Event Pattern Trial:} Participants were split into two groups, one using the positional patterns, and the other the semantic patterns. A simple graphical user interface (figure \ref{fig:ui}) was displayed on a laptop with labelled buttons representing each of the 8 patterns. During an initial training phase, participants would click these buttons to generate and feel the associated pattern. They had up to 10 minutes to learn the patterns (though most felt confident much earlier). After training, participants underwent a 60 second test repeated 6 times. Each test had event patterns generated in a random order at random intervals. Participants were instructed to click the corresponding button on the same GUI when they felt a message. In trials 4 and 5, the participants were also subjected to an additional mental load by tasking them to solve fast simple arithmetic problems out loud. Similarly, in the last trial, they were tasked to watch a tennis rally and count the times the ball crossed the vertical centre. These two sources of mental load were chosen to add to the user's cognitive load in the form of a problem solving, and a visual tracking task. We hypothesised that the semantic patterns were easier to learn and discern, expecting a stronger result from the group using them. The mental tasks were to simulate to some extent the additional cognitive duress that USAR operators might be subject to while interpreting information. 

\noindent \textbf{Direction and Movement Trial: } Participants were tasked to use a graphics tablet to draw the path they believed the robot was taking as it was teleoperated out of sight of the participant. The robot's path was saved by polling its odometry position, forming a ground truth with which the participant's drawings were compared. This trial was repeated 3 times for each participant, with different robot paths each time. This trial was identical for both participant groups.

\section{Results and Analysis}

18 participants took part in the study, primarily consisting of students from two local universities. Ethics approval was obtained through the University of Nottingham School of Computer Science. Before each trial, the vest was calibrated for the participant to ensure comfort. Each trial took 45 minutes on average.

\vspace{-8pt}
\subsection{Identification Accuracy and Selection Delay}

\emph{Identification Accuracy} is the ratio of correctly selected patterns. In general, the participants using the semantic patterns were significantly more accurate (\textit{p = 0.027}), with a mean accuracy of 79.18\%, compared to the positional patterns with 63.70\%. It was found that the additional mental tasks had no significant effect on the mean accuracy. 

\emph{Selection Delay} is the time between receiving a pattern, and the participant response. When completing the experiments without the additional mental tasks, there was no significant difference in delay for recognising the patterns, aside from more variance for the semantic group, likely due to the variety of the patterns. The additional mental tasks had a very significant (\textit{p=0.013}) effect on the positional group's selection delay, with participants taking on average an additional 0.5 seconds. This is not the case in the semantic group, where the mean is almost identical (1.71 and 1.79s respectively).

\begin{figure}
    \centering
    \begin{minipage}[b]{0.48\textwidth}
        \centering
        \includegraphics[width=0.9\textwidth]{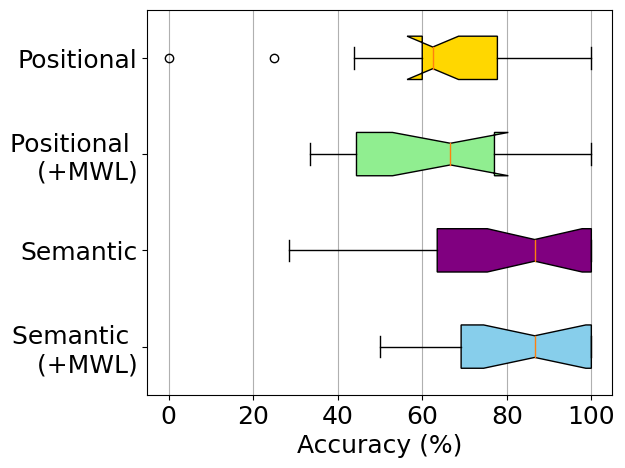}
        \caption{Identification Accuracy for both groups, with and without additional mental workload (MWL). The semantic pattern group performs significantly better}
        \label{fig:id_acc}
    \end{minipage}
    \hspace{0.02\textwidth}
    \begin{minipage}[b]{0.48\textwidth}
        \centering
        \includegraphics[width=0.9\textwidth]{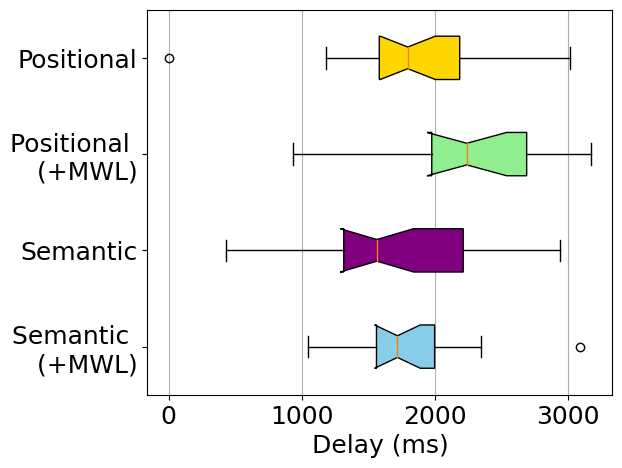}
        \caption{Selection delay for both groups, with and without additional mental workload. The positional group is significantly affected by the additional mental tasks.}
        \label{fig:id_del}
    \end{minipage}
\end{figure}

\subsection{Comparing Patterns}

\begin{figure*}
    \centering
    \begin{minipage}[b]{0.93\textwidth}
        \centering
        \includegraphics[width=1\textwidth]{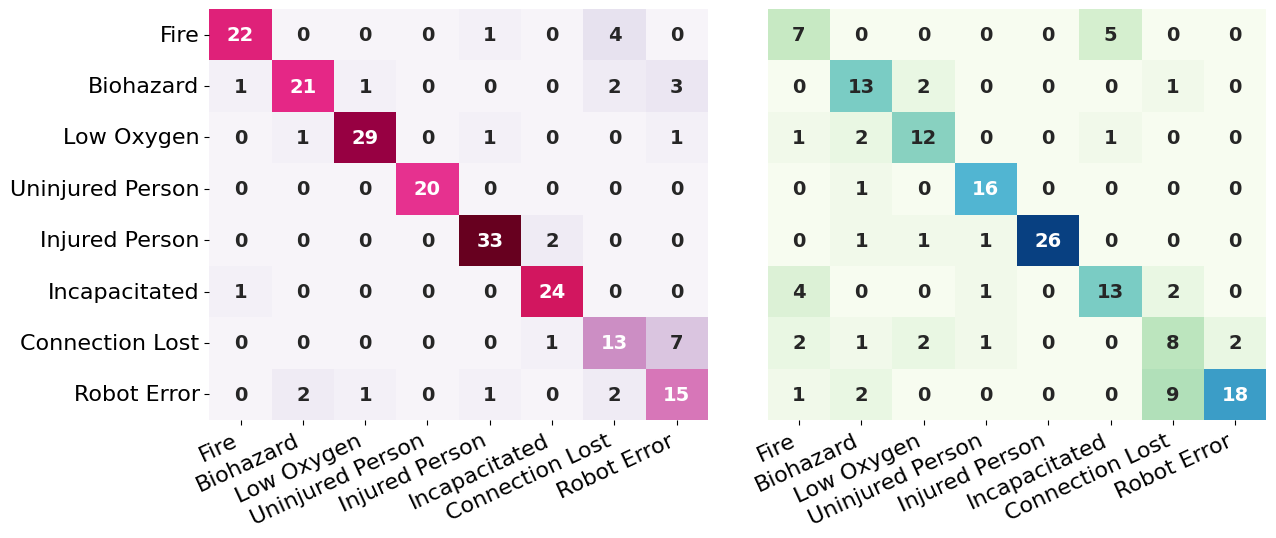}
    \end{minipage}
    \caption{Confusion Matrices for Patterns. Pink=semantic group, green=positional group. The semantic group performs slightly better, aside from the last two patterns.}
    \label{fig:g3}
\end{figure*}

The confusion matrices for both groups are shown in table \ref{fig:g3}. These show the semantic patterns were more discernible between each other. The confusion between \textit{Connection Lost} and \textit{Robot Error} messages in both groups was potentially due to the similar labels and images on the testing UI (figure \ref{fig:ui}).

\subsection{Path Similarity}

\begin{figure*}
    \centering
    \begin{minipage}[b]{0.34\textwidth}
        \centering
        \includegraphics[width=\textwidth]{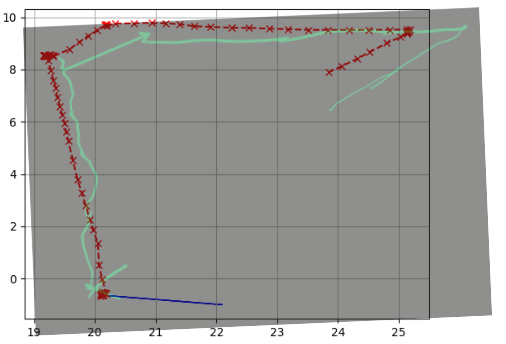}
        \label{fig:p1}
    \end{minipage}
    \begin{minipage}[b]{0.32\textwidth}
        \centering
        \includegraphics[width=\textwidth]{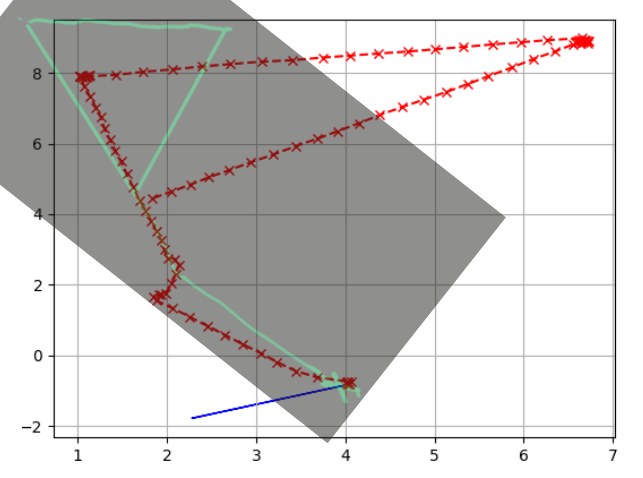}
        \label{fig:p2}
    \end{minipage}
        \begin{minipage}[b]{0.32\textwidth}
        \centering
        \includegraphics[width=\textwidth]{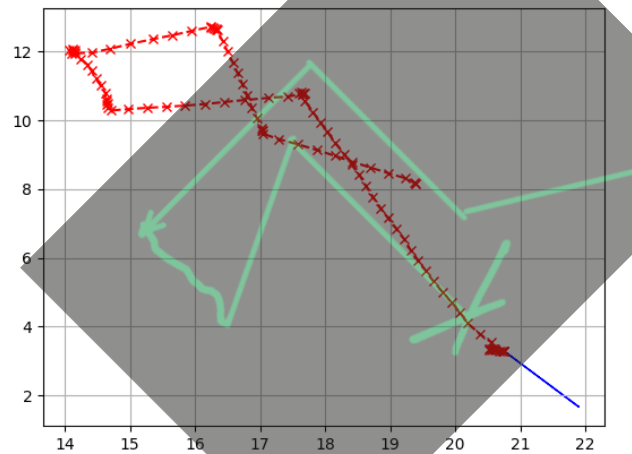}
        \label{fig:p3}
    \end{minipage}
    \caption{Example of robot paths (red) with participant interpretations (green). Participant drawings rotated to align with the path of the robot. Scale is in Meters.}
    \label{fig:pp3}
\end{figure*}

Due to technical issues involving the vest, and participant time constraints, 41 runs of this trial were completed (of a possible 54). In 31 of these (76\%), the participant correctly identified all turns the robot made, with the angle drawn within 20° of ground truth. In 23 of the trials (56\%), the participant identified the final position of the robot to within a metre. This analysis was conducted manually, by aligning and scaling the drawn path, then inspecting the similarities and differences between them. Examples of this can be found in figure \ref{fig:pp3}.

\vspace{-10pt}
\vspace{-13pt}

\subsection{Questionnaire Answers and Participant Feedback}

Table \ref{tab:data4} shows the responses to the questionnaire. Common feedback themes are discussed below. Participants are labelled \textit{A} for the semantic group, \textit{B} for the positional group.

\noindent \textbf{Learnability:} Participants in the semantic group expressed surprise at how easy the vibrohaptics were to learn, with some confident that with more time they could perform better. PA1: \textit{"... I think it’s really quick to pick up"},
PA3: \textit{"the patterns were designed correctly to be distinct ... it is very easy to learn and respond to them with little practice"},
PA8: \textit{"The event vibrations were easy to learn and mostly intuitive..."}.

\noindent \textbf{Difficulty Interpreting Lateral Orientations:} PA5:\textit{" when it was moving left or right it was hard to tell as there was no motors directly to the left and right"}. This is a downside of the TactVest X40. As it has no motors directly on the sides, one motor from each of the front and back parts of the vest was used to simulate that direction.

\noindent \textbf{Issues with Scale:} Participants found it difficult to associate the duration of vibration with the distance the robot travelled during the directional trial. PB1:\textit{"Understanding the distance the robot has moved was hard, if not impossible..."}. Whilst these patterns were never designed to give precise directions, this led to frustration. Interestingly, some participants reported feeling a change of motor intensity when the robot moved further away, but no such system was ever implemented or mentioned. 

\noindent \textbf{Participant Preferences:} Many participants noted areas of the vest where they struggled more to perceive the vibrations. Each participant had different preferences, leading to contradictory feedback. PA5: \textit{"Vibrations in back felt more accurate than on the front"},
PB6: \textit{"Actions displayed ... in the lower part of the backside of the vest were much harder to remember and respond to"}. This was also the case for feedback about some of the specific patterns. PA1: \textit{"Some were really easy such as the robot losing signal and error."}, PA8: \textit{"you could mistake the lost signal message for the robot error message..."}. This is a case for more in-depth calibration and customisation.

\begin{table}
    \caption{Mean response to questionnaire questions, split by group. Responses range from \textit{’Strongly Disagree’} (1) to \textit{’Strongly Agree`} (5).} 
    \begin{tabular}{p{0.6 \textwidth} p{0.20 \textwidth} p{0.20 \textwidth}}
        \hline
        \centering
        \textbf{Likert Measure} & \textbf{Positional} & \textbf{Semantic}\\
        \hline
        \textit{The vibrohaptics are Intuitive} & $3.3 \pm 0.83$ & $4.3 \pm 1.04$ \\
        
        \textit{The device is comfortable} & $5.0 \pm 0$ & $4.8 \pm 0.78$ \\
        
        \textit{The vibrations did not cause me discomfort} & $5.0 \pm 0$ & $4.2 \pm 1.08$ \\
        
        \textit{Noticing vibrations is easy} & $4.9 \pm 0.33$ & $5.0 \pm 0$ \\
        
        \textit{Locating vibrations is easy} & $4.3 \pm 0.97$ & $4.6 \pm 0.83$ \\
        
        \textit{The vibrations are distracting} & $2.4 \pm 1.58$ & $2.5 \pm 1.12 $ \\
        
        \textit{The event messages are easy to learn} & $2.8 \pm 1.21$ & $3.9 \pm 0.70$ \\
        
        \textit{The directional messages are easy to learn} & \multicolumn{2}{c}{$4.1 \pm 0.94$} \\
        \hline
    \end{tabular}
    \label{tab:data4}
\end{table}

\section{Discussion}

\subsection{Vibrotactile Messages for Information Delivery}

The results indicate that users find it intuitive and simple to associate vibrotactile sensations to certain events. The confusion matrices in figure \ref{fig:g3} show that even the positional patterns are still discernible from each other. 

Evaluating the directional messages is difficult, as it is hard to quantify how a person perceives the movements of the robot. The results from the conducted trial show that participants understood when and where the robot was moving, but struggled with scale. There were numerous comments that this form of communication was more distracting (as it was always on) than the event patterns. The approach tested in this study did not produce accurate results, and requires more refinement, or an overhaul. One such rework is mentioned in section \ref{future}.

\subsection{Semantic Vibrotactile Messages}    

One of the aims of this research was to assess if semantic vibrations are more intuitive, easy to learn and discernible than standard methods used in simple vibrotactile technology. The results found support our hypotheses. Participants who trialled the semantic patterns had better accuracy and identified messages faster, leading to the conclusion that these patterns took less effort to interpret, which is supported by the fact that additional mental load had minimal effect on accuracy and delay. Feedback from participants on semantic patterns was almost completely positive. The main implementation challenge is creating patterns that are distinct, noticeable, but also not distracting. This process was relatively simple in the case of only 8 different messages. Future research should assess if these findings are maintained when this number increases. Another observation is how semantic patterns were described as less comfortable. This observation reinforces the idea of customisation; users could configure and choose the patterns they find the most comfortable and effective.


\subsection{Limitations of this Research \& Future Directions}
\label{future}

Whilst almost all participants gave positive feedback for the system, these participants have no experience with USAR scenarios, are not rescuers, and all but one are in the 18-22 age range, which may be reflected in the results. Furthermore, due to space constraints it was impossible to test the capabilities of the vest with the participant in motion, or attempting strenuous physical activity, which obviously deviates from the real-life use case. Also the approach used to simulate additional mental workload associated with a typical USAR mission should be further evaluated, specifically assessing the users' cognitive load. It's also worth mentioning the user would be looking at a screen with buttons for the possible messages, which likely helped users identify messages. 

Future iterations of this research should take into account the opinions and feedback of rescuers, both in the development and testing stages. Alongside this, trials should mirror more realistic scenarios, incorporating motion and physical activity alongside more training time, with methods that slowly remove visual aids, until the user is able to discern the meaning of a pattern with no assistance.

Other significant research questions must also be considered. One challenge is engineering vibrotactile interfaces into Personal Protective Equipment (PPE) without compromising effectiveness, which may require new insights into motor placement and new semantic 'shapes'. Another promising research area is using the vest to guide users. Early informal tests showed positive results, successfully 'piloting' users to a location with the vest. This suggests that the vest could effectively provide 'directions' potentially outperforming voiced instructions, particularly in noisy environments. This could be displayed at the request of the user, granting the user a general direction, without constant distraction.

\section{Conclusions}

Overall, the study provides insights into using off-the-shelf commercial vibrotactile technology for unobtrusive information delivery, with trial findings providing some evidence for using the haptic channel as an intuitive interface. Our results show that users were able to discern and understand a variety of haptic patterns using a vibrotactile vest. This shows promise for a robot to convey real-time information using this modality, indicating that this can be effective, and importantly, intuitive. Whilst more testing is needed, these initial findings indicate that vibrotactile technologies can be used in a range of collaborative HRI scenarios where a user's vision and hearing may be occupied for other tasks. Another contribution of this research is the development of semantic vibrotactile patterns. By using the skin as if it were a screen, the semantic patterns we created were more memorable than traditional methods. Our results show that this approach has potential for richer vibrotactile feedback, however further research will be required for real-world application.

%
%
%
\bibliographystyle{splncs04.bst}
\bibliography{main}

\end{document}